\documentclass[journal,twoside,web]{ieeecolor}
\usepackage{generic}

\usepackage[utf8]{inputenc}

\usepackage{amsmath,amssymb,amsfonts}
\usepackage{textcomp}

\usepackage{array}
\usepackage{booktabs}
\usepackage{multirow}
\usepackage{graphicx}
\usepackage[table,xcdraw]{xcolor}
\usepackage{subfig}
\PassOptionsToPackage{normalem}{ulem}

\makeatletter

\pdfpageheight\paperheight
\pdfpagewidth\paperwidth

\providecommand{\tabularnewline}{\\}

\usepackage{times}
\usepackage{algpseudocode}

\makeatother

\def\BibTeX{{\rm B\kern-.05em{\sc i\kern-.025em b}\kern-.08em
		T\kern-.1667em\lower.7ex\hbox{E}\kern-.125emX}}
\begin{document}
%
\title{Beyond Fine-tuning: Classifying High Resolution Mammograms using Function-Preserving Transformations}
%

\author{Tao~Wei, 
	Angelica~I~Aviles-Rivero, 
	Shuo Wang, 
	Yuan Huang, 
	Fiona~J~Gilbert,\\  
	Carola-Bibiane~Sch\"{o}nlieb, 
	Chang~Wen~Chen
\thanks{This work was supported in part by the CMIH, University of Cambridge. EPSRC grant Nr. EP/M00483X/1, the EPSRC Centre Nr. EP/N014588/1.}%
\thanks{T Wei and CW Chen are with the Department 
of Computer Science, State University of New York at Buffalo, New York, USA. \{taowei,chencw\}@buffalo.edu}
\thanks{AI Aviles-Rivero is with the Department of Pure Mathematics and Mathematical Statistics, University of Cambridge, UK  ai323@cam.ac.uk. }
\thanks{S Wang is with the Data Science Institute, Imperial College London, UK. shuo.wang@imperial.ac.uk}
\thanks{Y Huang and FJ Gilbert are with the Department of Radiology, University of Cambridge, UK. \{yh288,fjg28\}@cam.ac.uk}
\thanks{CB Sch\"{o}nlieb is with the Department  of Applied Mathematics and Theoretical Physics, University of Cambridge, UK. cbs31@cam.ac.uk}
}

\maketitle

\begin{abstract}
	The task of classifying mammograms is very challenging because the lesion  is usually small in the high resolution  image. The current state-of-the-art approaches for medical image classification rely on using the de-facto method for ConvNets - fine-tuning. However, there are fundamental differences between natural images and medical images, which based on existing evidence from the literature, limits the overall performance gain when designed with algorithmic approaches. In this paper, we propose to go beyond fine-tuning by introducing a novel framework called MorphHR, in which we highlight a new transfer learning scheme. The idea behind the proposed framework is to integrate function-preserving transformations, for  any continuous non-linear activation neurons, to internally regularise the network for improving  mammograms classification.
	The proposed solution offers two major advantages over the existing techniques. Firstly and unlike fine-tuning, the proposed approach allows for modifying not only the last few layers but also several of the first ones on a deep ConvNet. By doing this,  we can design the network front to be suitable for learning domain specific features. Secondly, \textcolor{black}{the proposed scheme is scalable to hardware}. Therefore, one can fit high resolution images on standard GPU memory. We show that by using high resolution images, one prevents losing relevant information. We demonstrate, through numerical and visual experiments, that the proposed approach yields to a significant improvement in the classification performance over state-of-the-art techniques, and is indeed on a par with radiology experts. Moreover and for generalisation purposes, we show the effectiveness of the proposed learning scheme on another large dataset, the ChestX-ray14, surpassing current state-of-the-art techniques.

\end{abstract}

\begin{IEEEkeywords}
	Deep Learning, Transfer Learning, Mammogram Classification, High Resolution, Network Morphism, Function-Preserving Transformations.
\end{IEEEkeywords}

%
\IEEEpeerreviewmaketitle

%
%
%

\section{Introduction}



%
%
%
%


\IEEEPARstart{S}{creening} mammography is the primary imaging test for early detection of breast cancer as it offers a reliable and reproducible test for diagnosis. However, a major challenge is the interpretation of the mammogram, which requires specialised training and substantial experience by the
reader~\cite{LehmanAraoEtAl2016,becker2017deep}. 
Mammograms can be difficult to read due to high variability in the patterns and subtle appearances of small cancers. This can result in variation in performance between experts~\cite{gilbert2006single,elmore2009variability,lehman2015diagnostic}. As a result, it is often necessary to advocate to double reading of mammograms at the expense of increasing the cost and expert workload.

The aforementioned drawbacks have motivated the rapid development of automatic and robust algorithmic approaches to support the experts' outcome. In particular, computer-aided systems, that aim to influence the expert interpretation, have shown limited performance~\cite{LehmanWellmanEtAl2015}. This limitation is mainly because traditional systems are designed using hand-crafted features~\cite{warren1995comparison}.
Most recently, with the astonishing success of deep learning (DL), it has been shown that these systems can substantially improve their performance by learning features as data representatives~\cite{HamidinekooDentonEtAl2018}.

At the algorithmic level, breast diagnosis can be casted as a classification task, in which several developments using deep ConvNets have been reported e.g.~\cite{huynh2016digital,levy2016breast,geras2017high,shen2019deep}.  However despite the rapid development of DL based techniques,  mammography classification  remains an unsolved problem. Therefore, the question of how to improve the classification performance is of great interest from the technical and clinical points of view and is the problem that we address in this work.

There have been different attempts to design end-to-end solutions, in which the majority are based on pre-trained and fine-tuned~\cite{KrizhevskySutskeverHinton2012,HeZhangRenEtAl2015} using for example ResNet~\cite{he2016deep} type architectures. \textcolor{black}{However, one of the major challenges in mammography classification is that {\it mammograms are high resolved in nature} and 
{\it the abnormality is significantly smaller than the whole mammogram image}, for example, $100\times 100$ vs $3$k $\times$ $5$k. The majority of existing algorithmic approaches are usually not scalable to hardware. They cannot directly handle the high resolution nature of medical images and are designed to work around the limits of the GPU memory} by either: i) decreasing significantly the image resolution or ii) to use image patches for the classifier.  Although these two alternatives have shown promising results, they are not ideal as they compromise the clinical outcome. When the resolution is decreased, the negative effects are reflected at the level of losing relevant small clinical regions; for example, the structures might appear blurred. Whilst when a patch-based solution is designed, it might suffer from significant boundary artefacts~\cite{InnamoratiRitschelEtAl2018}. 

In this work, we propose a novel learning scheme, that we called MorphHR, that addresses the current drawbacks in the existing literature.  We show that the proposed approach significantly boosts the classification accuracy whilst demanding low GPU usage. Our main contributions are as follows: 

\begin{itemize}
	\item \textcolor{black}{We reveal that resolution is critical for mammogram classification. It is intuitive that mammograms are high resolved ($\sim$4K) in nature. However, due to the limit of GPUs memory, current mainstream approaches either decrease the resolution or extract patches. We argue that the importance of resolution for medical image classification should be emphasised. In particular, this work is proposed to resolve this resolution issue.}
	\item  We propose a new learning scheme, which uses function-preserving transformation to regularise the network. In particular, we build the proposed framework using 
	Network Morphism~\cite{WeiWangRuiEtAl2016} principles
	as proxy task for transferring knowledge from different domains, specially, from low resolution natural images to high resolution mammograms. This leads to the next advantages.
	\begin{itemize}
		\item By introducing this proxy task in the training, we can go beyond fine-tuning by modifying not only the last layers but also the first few ones of the deep Net. This will cater to its own domain feature learning.
		\item The proposed training scheme \textcolor{black}{is {\it scalable} to hardware} and allows for using high resolution mammograms that fit standard GPU memory. This will avoid losing relevant clinical regions - as the abnormalities are very small.
	\end{itemize}
	\item We extensively evaluate the proposed approach on the CBIS-DDSM~\cite{LeeGimenezEtAl2017} dataset using an extensive numerical and visual experiments. Moreover and for generalisation purposes, we show how the transfer knowledge can be effective on other large datasets such as the ChestX-ray14~\cite{WangPengEtAl2017}.
	\item We show that the proposed learning scheme mitigates the current limitations of the body of literature, by outperforming current state-of-the-art techniques in mammography, which is on a par with radiology experts, and X-ray classification. 
\end{itemize}

\section{Related Work}
The problem of classifying mammography data has been widely investigated in the community, in which the solutions are based on using hand-crafted or automatic selected features. In this section, we review the body of literature in turn. We then remark the current drawbacks and motivate the proposed novel learning scheme.  \smallskip

There have been different attempts to deal with the mammography classification task. Early developments were limited by their own construction, as they were based on the use of hand-crafted features such as texture analysis or intensity-based reasoning e.g. ~\cite{brzakovic1990approach,petrosian1994computer,vyborny1994computer}. These algorithmic approaches were based on strict modelling hypothesis such as conditioning the intensity histogram e.g.~\cite{brzakovic1990approach} - therefore, parameters adjustments were necessary for every single image. Therefore, these methodologies were not robust and generalisable to sightly changes across them.

The previous drawbacks were mitigated with the remarkable success of deep learning in computer vision; in tasks such as object recognition and detection~\cite{KrizhevskySutskeverHinton2012,HeZhangRenEtAl2015}. This has motivated the community to apply successfully DL techniques to medical data.  However, this type of data has set new challenges as significantly different from natural images. For example, the low signal-to-noise ratios or small relevant anatomical structures  need to be taken into account to avoid false positive or false negative outcomes.

In particular, for the task of mammography data classification, there have been different works reported. The mainstream approach for this problem has been the use of patch-based classifiers e.g.~\cite{XiShuEtAl2018,RampunWangEtAl2018,mercan2017multi,AgarwalDiazEtAl2019,Chun-mingXiao-meiEtAl2019,RagabSharkasEtAl2019,wu2019deep}.
The central idea, of this perspective, is to discompose the mammograms in image patches which reflect: i) areas with abnormalities (positive patches) and ii) normal regions (negative patches).
Although these algorithmic approaches have reported promising results, when a patch-based approach is applied to a whole image, one can observe significant boundary artefacts. \cite{InnamoratiRitschelEtAl2018}

A key reference  in this category is the work of Li Shen~\cite{Shen2017,ShenMargoliesEtAl2019}, which to the best of our knowledge holds the state-of-the-art results on the CBIS-DDSM dataset~\cite{LeeGimenezEtAl2017}. The authors' central idea is to use an all convolutional design to convert a patch-based classifier to an image-based one. However, the approach is not scalable to the image input size. That is -  when the image size doubles both the computational run-time and memory shall require 4x resources. The computational resources will then overflow and reach their capacity. 
Further, a pre-trained patch-classifier heavily rely on having region-of-interests (ROIs) annotations, which is expensive and laborious to create.

Another set of algorithmic approaches have addressed the mammography problem as the task of detecting \cite{Akselrod-BallinKarlinskyEtAl2016, RibliHorvathEtAl2018, AgarwalDiazEtAl2019} or segmenting \cite{RonnebergerFischerBrox2015, SunLiEtAl2018} the abnormal regions. However, the main drawback of these perspectives is that they require 
finer annotations including ROI boxes or contours \cite{ShenMargoliesEtAl2019}.
These annotations are expensive to collect and are generally unavailable in the datasets.

Although, the aforementioned approaches have reported promising results, they are limited by their own construction. The limitation comes from a commonality: the use of fine-tuning. That is,  a DL model is first trained on a large dataset, such as ImageNet~\cite{deng2009imagenet}, and then it is fine-tuned for another task. However, the limitation of fine-tuning is that one can only modify the last several layers in a deep net. It is then not possible to alter the front layers. This limitation has motivated our current work. We propose a learning scheme that allows modifying the first several layers of any deep net, this, with the goal of \textit{ transferring knowledge from natural images to high resolution medical images without requiring high GPU memory}. 

\textcolor{black}{The proposed framework is inspired by the ideas of using function-preserving transformations, i.e., Network Morphism~\cite{WeiWangRuiEtAl2016}, in which one seeks to transfer the deep nets knowledge effectively. However, we further clarify the difference between that work and ours. Firstly,  we use the principles from ~\cite{WeiWangRuiEtAl2016} as proxy task as part of the proposed framework. Secondly, unlike ~\cite{WeiWangRuiEtAl2016}, our goal is to improve the mammograms classification performance. Thirdly, we carefully design the morphing operations and strides to handle the resolution problem for mammogram classification. While in~\cite{WeiWangRuiEtAl2016}, the authors carry out the morphing operations on the same level without resolution changes.}



\section{Proposed Approach}
This section contains two main part: i) the differences between fine-tuning and our philosophy and  ii)  the proposed framework for classifying mammography data. In what follows, we first start formalising our problem.

\medskip
\textbf{Problem Definition} Giving a training set in the form of pairs $\{(x_1,y_1),...,(x_n,y_n)\}$, with input-target attributes $x_i$ and $y_i$ correspondingly. We seek to find a optimal classifier $f: X\rightarrow Y$ 
that maps the input  $X$ to the output $Y$ space, with minimum generalisation error, such that $f$ works well with unseen instances.

\subsection{Beyond Fine-Tuning: Morphing Deep Nets for Mammography Data}

In this section, we underline the need for improving upon fine-tuning. We highlight the major drawbacks of it when using medical images and give initial insights into the proposed approach.
\smallskip

Deep neural networks usually have millions of parameters and require
a significant amount of data samples to train, whereas mammogram images
are usually expensive to collect and the publicly available datasets
are smaller than is needed for training. To address the limited-data 
problem, a technique to-go is fine-tuning. The key idea of it is to train a deep Net on a 
reasonably large dataset such as the ImageNet \cite{deng2009imagenet} dataset; and then the last layer (or last few layers) of the pre-trained neural network are dropped
and replaced with a new layer (or several layers) to be fine-tuned
for classification on a new task. 


\begin{figure}[t!]
	\begin{centering}
		\includegraphics[width=0.9\linewidth]{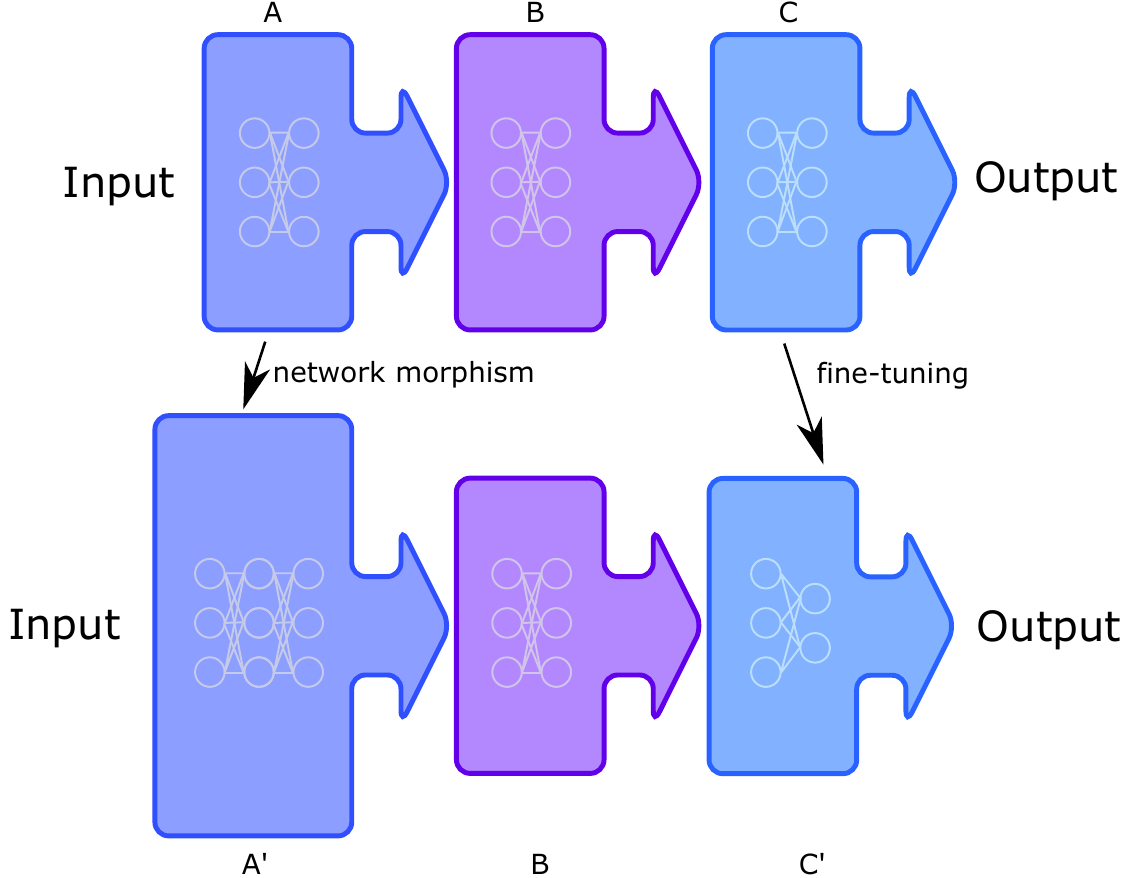} 
		\par\end{centering}
	\caption{Network morphism vs fine-tuning for knowledge transferring. Fine-tuning is the de-facto method for deep learning applications to medical imaging. However, it only allows to alter the last few layers ($C\to C'$) of a neural network. In this research, we propose to use network morphism to alter the first few layers ($A\to A'$) of a neural network to	help to transfer knowledge from the natural images domain to the medical images domain.\label{fig:concept}}
\end{figure}

Fine-tuning is the de-facto method for DL applications
to medical imaging. However, there are fundamental differences in
data sizes, features, and resolutions between the natural image domain and the medical
image domain. Deep ConvNets are usually designed for natural image tasks.
Therefore, the question of -  how much of the ImageNet (or other natural images datasets) feature reuse is helpful for medical images. This question has been addressed in several works. For example, the authors of that~\cite{RaghuZhangEtAl2019} showed that fine-tuning transfer could offer little benefit
to medical classification performance. In \cite{KornblithShlensEtAl2019},
it was illustrated that pre-trained features may be less general than
previously thought. \textcolor{black}{Besides, fine-tuning is not scalable to the resolution of the input image. As illustrated in Fig.~\ref{fig:concept}, suppose that the input image resolution is doubled, all convolutional layers of $A$, $B$, $C$ shall require 4x GPU FLOPs and memory. Hence, current mainstream approaches based on fine-tuning, using either reduced resolutions or patches, are limited in the exploration of full high resolved mammograms.}

\begin{figure*}[t!]
	\begin{centering}
		\includegraphics[width=1\linewidth]{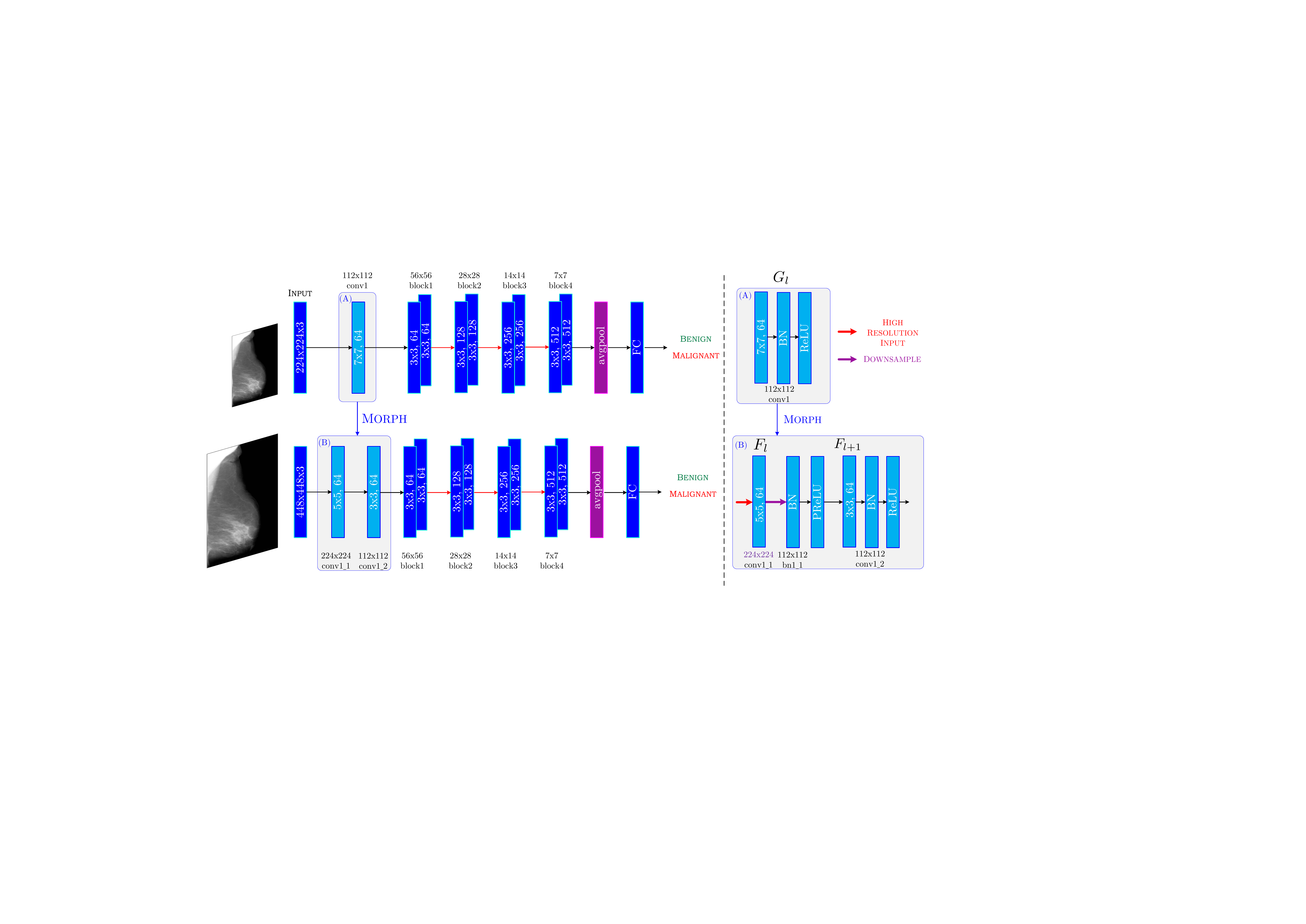}
		\par\end{centering}
	\caption{Illustration of the proposed MorphHR scheme. The first convolutional
		layer with kernel size of $7\times7$ is morphed into two convolutional
		layers with kernel sizes of $5\times5$ and $3\times3$. The receptive
		field of the morphed network is kept and the network function is preserved.
		We double the input image size and set the stride of \texttt{conv1\_1} to be
		2 for downsampling. The layers in \texttt{block1} and after are exactly the same. The morphed network is further fine-tuned on the mammography data.} \label{fig:illustration}
\end{figure*}

Therefore, there is a need to mitigate the fine-tuning limitations in the medical domain, \textcolor{black}{allowing to be scalable to high resolved inputs}. It is important to modify not only the last layers but also the first few ones of a deep Net. This to cater for the own domain feature learning. To this aim, we propose a novel learning scheme that allows morphing deep nets without altering the function of the backbone network, it builds upon the transformation principles from~\cite{WeiWangRuiEtAl2016}. The major differences between fine-tuning and the adopted learning scheme is illustrated in Fig.~\ref{fig:concept}. In that figure, one can alter the front of a neural network from $A$ to $A'$.  \textit{The first major benefit of this operation is that we can design the block $A'$ to be suitable for learning domain specific features.}


Whilst \textit{the second major benefit of the proposed scheme is that it 
allows for using high resolution mammograms that fit in a standard GPU memory.} Why does it work with highly resolved images? \textcolor{black}{As illustrated in Fig.~\ref{fig:concept}, suppose that the input image resolution is doubled, we are able to carefully design the morphing front $A'$ such that $B'$ and $C'$ maintain their original GPU FLOPs and memory consumption.} Deep ConvNets are typically designed to fit images with input size of, for example, $224\times224$ or $227\times227$. This input image size is fixed
until one modifies the network architecture by: i) replacing fully
connected layers with $1\times1$ convolutional layers or ii) replacing
the last pooling layer with a global/adaptive pooling layer.  However, these
two approaches have limitations: a) they fail if there is a
single layer that requires fixed input size and it can not be modified
to accept other input sizes; and b) they are not scalable,
i.e., when the image size double both the computational run-time and
memory shall require 4x resources.

We mitigate the existing aforementioned drawbacks by introducing a learning scheme that allows morphing deep Nets - it allows for using high resolution images whilst being scalable. We give precise details of the proposed algorithmic approach in the next section.

\subsection{MorphHR: Classifying Mammography Data}
We propose a novel framework for classifying mammography data, which is capable of transferring knowledge between natural and medical images. Moreover, it works on high resolved images which fit on standard GPU memory and is scalable. In what follows, we give details of the proposed approach, which involves function-preserving transformations. 


In this work, we adopt ResNet18\footnote{In the proposed experiments, RestNet50 did not boost the performance significantly over ResNet18 on the mammography data.  In \cite{RaghuZhangEtAl2019}, the authors showed similar findings, that is - smaller architectures can perform comparably to standard ImageNet models. Hence, we selected ResNet18.} as the backbone for our framework. However, the proposed scheme is generic and one can use other deep Nets architectures as long as the first layer is a convolutional layer. The first problem to deal with is the high resolution of the mammograms - typically 3k$\times$5k pixels, which are significantly larger than the resolution of natural images. To mitigate this problem, we propose to morph the first layers of our backbone such that it allows for high resolved images whilst requiring standard GPU memory.

\textbf{MorphHR Design.}
The proposed approach, MorphHR, is illustrated in Fig.~\ref{fig:illustration}. One can observe that the  \texttt{conv1} is morphed into two convolutional layers \texttt{conv1\_1} and \texttt{conv1\_2}. \textit{The added layer allows  more precisely learning on the features of the mammographic images.} While primitive
features, such as edges, corners, and textures, are universal for images, they can however be considerably different.  By modifying the front layer, one can adapt these primitive features more efficiently. This modification on the backbone also allows the convolutional kernel size of \texttt{conv1\_1} to be $5\times5$ and  \texttt{conv1\_2} to be $3\times3$. Hence, the effective receptive field size is $5+3-1=7$, which is equal to
the original convolutional kernel size $7\times7$ and the receptive
field size is kept.

Besides the added parameters at the front of the neural network to facilitate learning, we
also change the size the image input in order to promote the resolution.
We double the image input size and also insert a maxpool layer of
kernel size $2\times2$ after \texttt{conv1\_1}. It is easy to verify that the feature
output dimensions of layer \texttt{conv1} and \texttt{conv1\_2} are both of shape (64,112,112).
Hence, the computation of layers \texttt{conv1}/\texttt{conv1\_2}, and the ones after, are preserved.
At the implementation level, the network function shall be perturbed when we double the image input size and half the feature output. However, ConvNets are robust enough
to handle such small perturbations. In the proposed experiments, the network recovers to the original accuracy right after several iterations. In summary, the
above operation allows us to promote the image input size on a pre-trained
neural network without significantly altering the overall network
architecture. This operation can be applied
for a second round to promote the image input size to be 896.

\textbf{How to Morph Deep Nets? Function-Preserving Transformations.} Mathematically, to morph the deep net displayed in Fig.~\ref{fig:illustration}, to a new one with the network function completely preserved, one needs to compute the convolution operation as follows:

\begin{equation}
O_{j}(c_{j})=\sum_{c_{i}}O_{i}(c_{i})*G_{l}(c_{j},c_{i}),\label{eq:conv}
\end{equation}
where the output blobs $O_{*}$ are 3D tensors of shape $(C_{*},H_{*},W_{*})$
and the convolutional filter $G_{l}$ is a 4D tensor of shape $(C_{j},C_{i},K_{l},K_{l}).$
We define $C_{*}$, $H_{*}$, and $W_{*}$ as the number
of channels, height and width of $O_{*}$ correspondingly; and $K_{l}$ is the convolutional
kernel size. 

The convolutional filter $G_{l}$ is morphed into two convolutional
filters $F_{l}$ and $F_{l+1}$ (Fig. \ref{fig:illustration} right),
where $F_{l}$ and $F_{l+1}$ are 4D tensors of shapes
$(C_{l},C_{i},K_{1},K_{1})$ and $(C_{j},C_{l},K_{2},K_{2})$. In order to preserve the network function,
one needs to morph the network, which expression reads: 

\begin{equation}
\tilde{G_{l}}(c_{j},c_{i})=\sum_{c_{l}}F_{l}(c_{l},c_{i})*F_{l+1}(c_{j},c_{l}),\label{eq:morph}
\end{equation}
where $\tilde{G_{l}}$ is a zero-padded version of $G_{l}$ whose
effective kernel size is $\tilde{K_{l}}=K_{1}+K_{2}-1\geq K_{l}$. The sufficient condition to morph a network was shown in~\cite{WeiWangRuiEtAl2016}, we therefore seek to fulfill the next condition:

\begin{equation}
\max(C_{l}C_{i}K_{1}^{2},C_{j}C_{l}K_{2}^{2})\geq C_{j}C_{l}(K_{1}+K_{2}-1)^{2},\label{eq:exact_condition}
\end{equation}

\noindent
in the proposed case illustrated in Fig. \ref{fig:illustration}, we have $C_i=3,  C_j=64, C_l=64, K_1=5, K_2=3$. From \eqref{eq:exact_condition} one can see that $\max(4800, 36864) \geq 9408$ holds. 

Besides convolutional layers, the BatchNorm layers and ReLU layers needs to be carefully addressed. We adopt the approach of that~\cite{WeiWangChen2019} for BatchNorm layers by setting: \texttt{bn1\_2} as \texttt{bn1}; and \texttt{bn1\_1} by using  $\gamma=1$ and $\beta=0$. We refer to  $\gamma$ and $\beta$ as the  parameters of a BatchNorm layer \cite{IoffeSzegedy2015}. For the non-linear activation layer ReLU, we adopt the approach in \cite{WeiWangRuiEtAl2016} by using a proxy version of PReLU with the slope set to 1. 

\begin{figure*}
	\begin{centering}
		\subfloat[ResNet18\_S224]{\begin{centering}
				\includegraphics[width=0.33\linewidth]{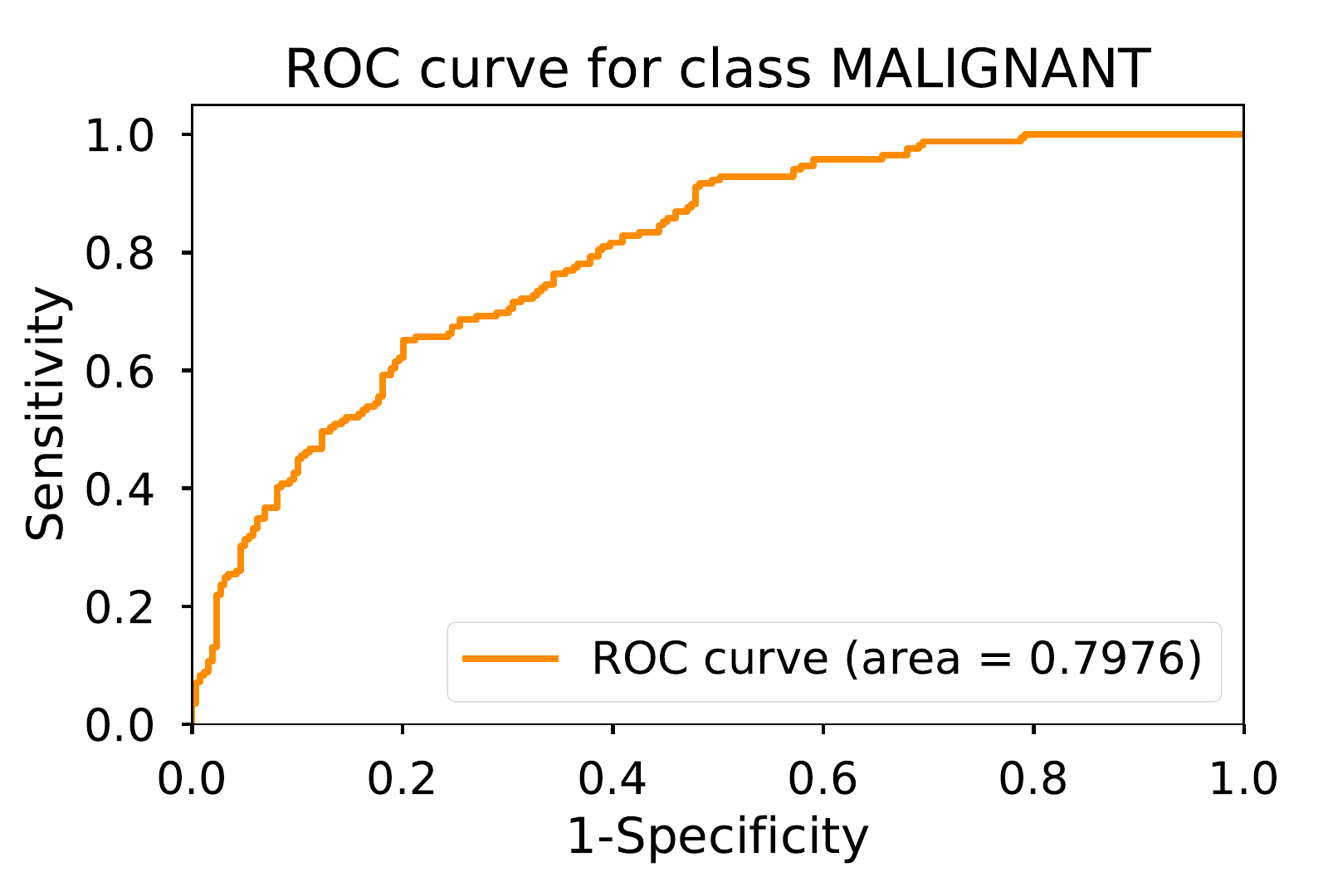} 
				\par\end{centering}
		}\subfloat[ResNet18\_S448]{\begin{centering}
				\includegraphics[width=0.33\linewidth]{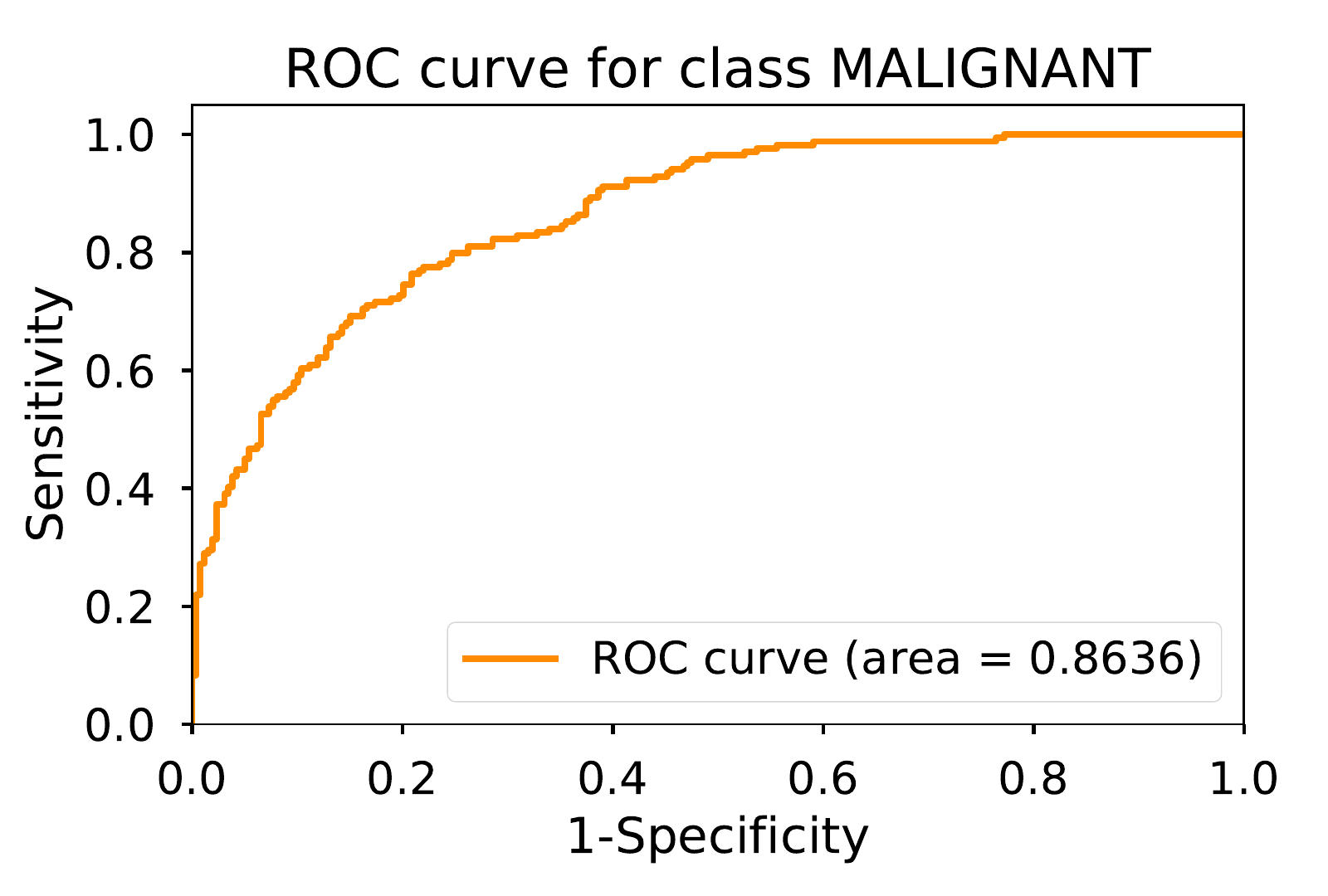} 
				\par\end{centering}
		}\subfloat[ResNet18\_S896]{\begin{centering}
				\includegraphics[width=0.33\linewidth]{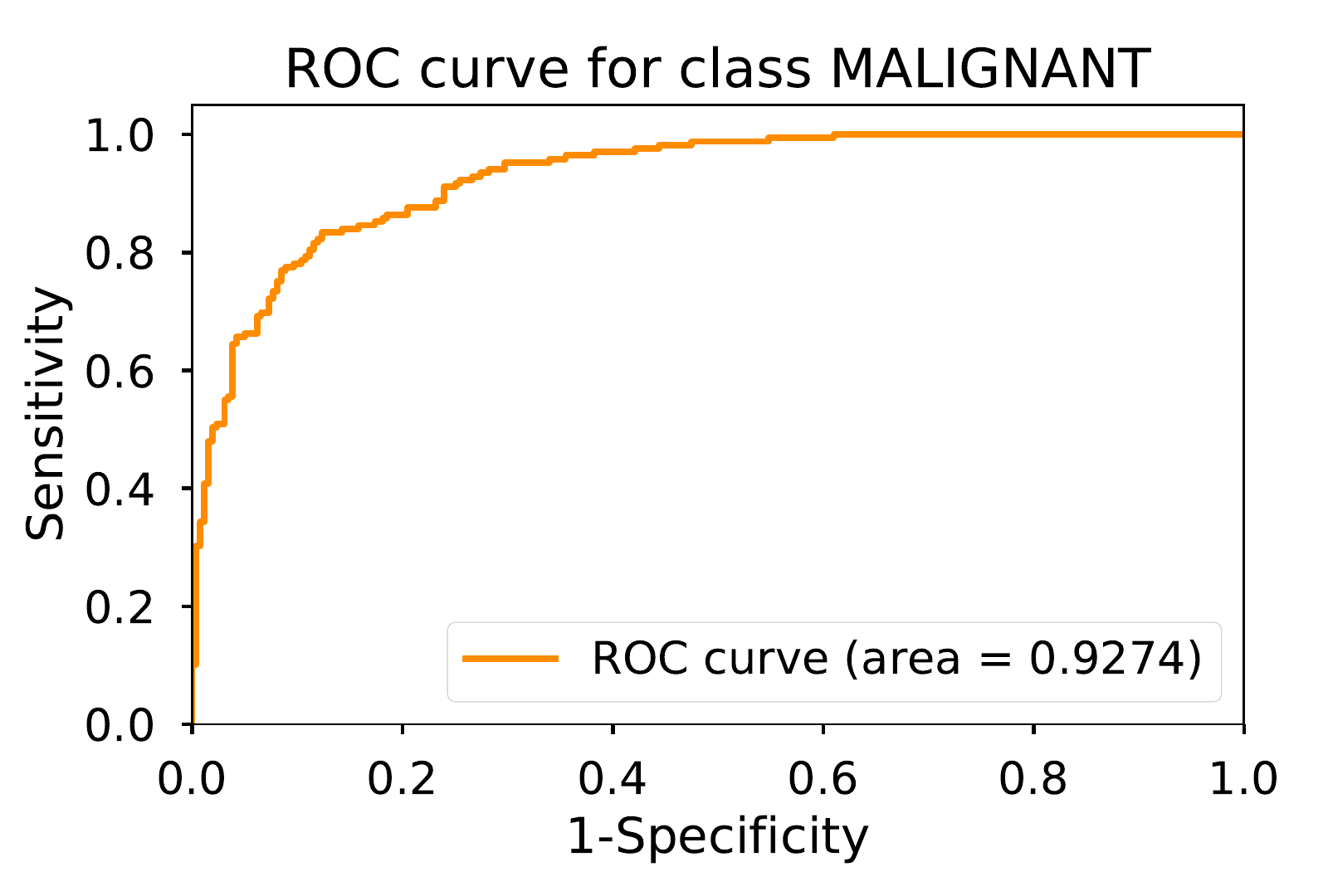} 
				\par\end{centering}
		}
		\par\end{centering}
	\vspace*{-10pt}
	\begin{centering}
		\subfloat[MorphHR-ResNet18\_S224]{\begin{centering}
				\includegraphics[width=0.33\linewidth]{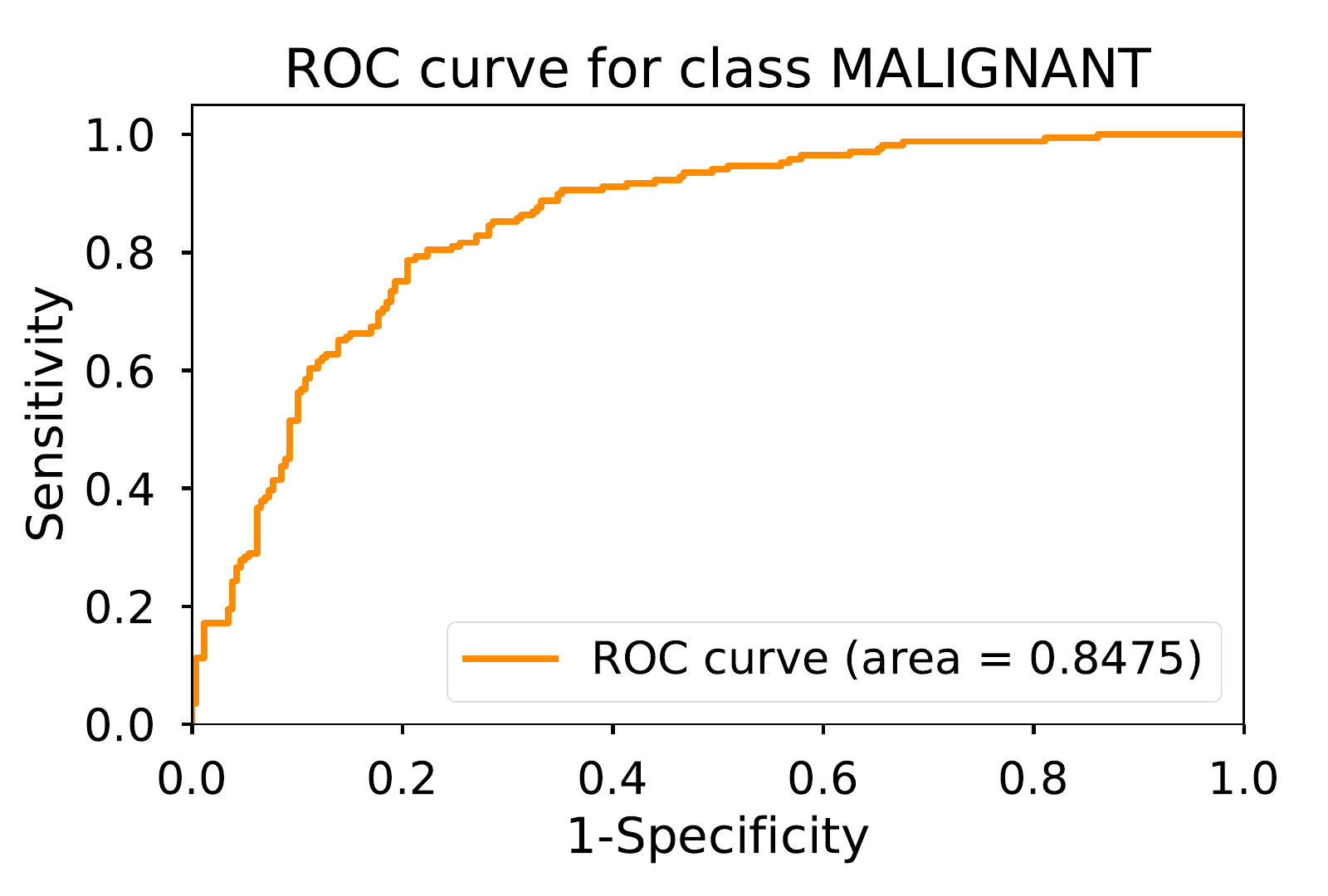} 
				\par\end{centering}
		}\subfloat[MorphHR-ResNet18\_S448]{\begin{centering}
				\includegraphics[width=0.33\linewidth]{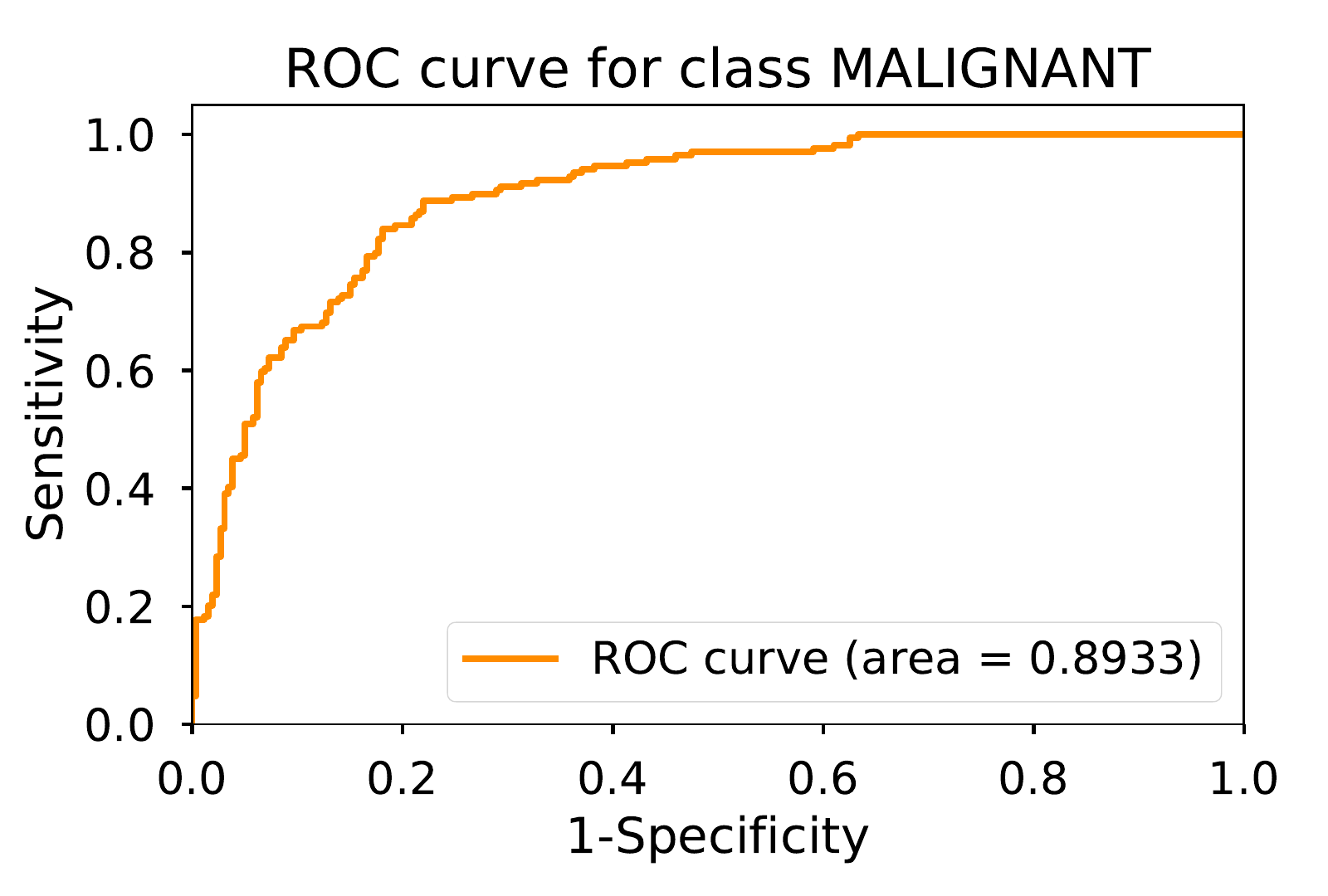} 
				\par\end{centering}
		}\subfloat[MorphHR-ResNet18\_S896]{\begin{centering}
				\includegraphics[width=0.33\linewidth]{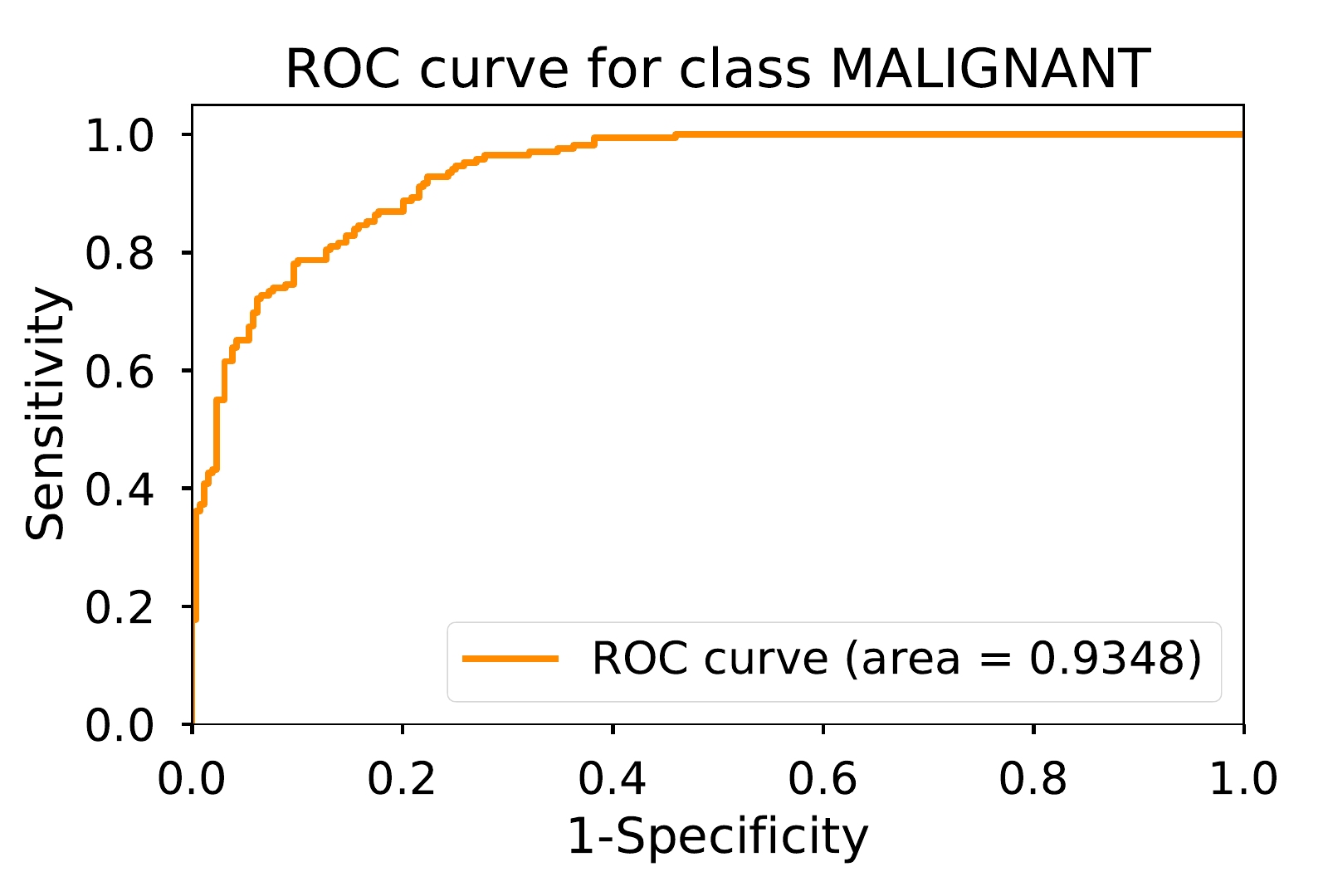} 
				\par\end{centering}
		}
		\par\end{centering}
	\caption{ROC curves on the CBIS-DDSM dataset. Image resolution plays a critical role in the task of screening mammography classification. \label{fig:ddsm_roc_curve}}
\end{figure*}

\begin{table*}
	\begin{centering}
		\begin{tabular}{ccccccc}
			\toprule 
			\multirow{1}{*}{} & \cellcolor[HTML]{EFEFEF} Layer4  & \cellcolor[HTML]{EFEFEF} Input  & \cellcolor[HTML]{EFEFEF} Val  & \cellcolor[HTML]{EFEFEF} Val  & \cellcolor[HTML]{EFEFEF} Test  & \cellcolor[HTML]{EFEFEF} Test\tabularnewline
			& \cellcolor[HTML]{EFEFEF} Computed on  & \cellcolor[HTML]{EFEFEF} Size  & \cellcolor[HTML]{EFEFEF} (1-crop)  & \cellcolor[HTML]{EFEFEF}(10-crops)  &\cellcolor[HTML]{EFEFEF} (1-crop)  & \cellcolor[HTML]{EFEFEF} (10-crops)\tabularnewline
			\midrule
			\midrule 
			ResNet18\_S224  & 7x9  & 224x288  & 78.15  & 79.76  & 72.57  & 75.33\tabularnewline
			\midrule 
			MorphHR-ResNet18\_S224  & 7x9  & 448x576  & \cellcolor[HTML]{DCEDC8} 82.5  & \cellcolor[HTML]{DCEDC8} 84.75  & \cellcolor[HTML]{DCEDC8} 75.23  & \cellcolor[HTML]{DCEDC8}76.73\tabularnewline
			\midrule 
			\midrule  
			ResNet18\_S448  & 14x18  & 448x576  & 83.82  & 86.36  & \cellcolor[HTML]{DCEDC8} 78.82  & 79.95\tabularnewline
			\midrule 
			MorphHR-ResNet18\_S448  & 14x18  & 896x1152  & \cellcolor[HTML]{DCEDC8} 87.48  & \cellcolor[HTML]{DCEDC8}89.33  & 78.36  & \cellcolor[HTML]{DCEDC8}80.13\tabularnewline
			\midrule 
			\midrule  
			ResNet18\_S896  & 28x36  & 896x1152  & 91.26  & 92.74  & 79.58  & 80.68\tabularnewline
			\midrule 
			MorphHR-ResNet18\_S896  & 28x36  & 1792x2304  & \cellcolor[HTML]{DCEDC8} 91.82  & \cellcolor[HTML]{DCEDC8} 93.48  & \cellcolor[HTML]{DCEDC8} 79.64  & \cellcolor[HTML]{DCEDC8} 81.87\tabularnewline
			\midrule 
			MorphHR-ResNet18\_S896 (Ensemble)  & 28x36  & 1792x2304  & \cellcolor[HTML]{DCEDC8} 93.60  & \cellcolor[HTML]{DCEDC8} \textbf{94.27}  & \cellcolor[HTML]{DCEDC8}82.16  & \cellcolor[HTML]{DCEDC8} \textbf{83.13}\tabularnewline
			\bottomrule
		\end{tabular}
		\par\end{centering}
	\caption{Experimental results of the proposed MorphHR scheme on the
		CBIS-DDSM dataset. \label{tab:ddsm_results}}
\end{table*}

\textbf{Morphing Deep Nets for Classification.} For the mammogram classification task, we use the cross entropy loss function to train our neural network. For mammography data, there
are only two classes, either a benign tumour or a malignant tumour.
However, the number of malignant tumor cases are usually significantly
less comparing against the benign cases. This imposes a significant class-imbalance
problem. Hence, we use the following weighted version of the cross
entropy loss, which reads:
\begin{alignat}{1}
l(o_{i},y_{i})= & -w[y_{i}]\left(\log\left(\frac{\exp(o_{i}[y_{i}])}{\sum_{j}\exp(o_i[j])}\right)\right)\\
= & w[y_{i}]\left(-o_{i}[y_{i}]+\log\left(\sum_{j}\exp(o_{i}[j])\right)\right),\label{eq:loss_wce}
\end{alignat}
\noindent
where $o_{i}$ is the predicted output for input image $x_{i}$ whilst
$y_{i}$  the class label of $x_{i}$. Moreover, $w$ refers to the weight vector
for all the classes. We set $w[y_{i}]$ to be inverse proportional
to the number of cases in class $y_{i}$, and the mean equals
to 1. That is:
\begin{equation}
w[y_{i}]=\frac{C}{\#cases[y_{i}]}/\left(\sum_{j}\frac{1}{\#cases[j]}\right),
\end{equation}
where $C$ is the total number of classes, $\#cases[y_{i}]$ is the
number of cases in the training dataset for class $y_{i}$. With this setup, in one training epoch, all the classes have equal weight of contribution to the loss function. Experimental results show that this weighted cross entropy loss is critical to the performance.




\section{Experimental Results}
In this section, we detail the experiments carried out to evaluate the proposed MorpHR scheme.

\subsection{Datasets Description}
We use the benchmarking  CBIS-DDSM \cite{LeeGimenezEtAl2017} dataset to evaluate the proposed approach. It is composed of 3103 mammography images from 1566 women. For each breast, both
craniocaudal (CC) and mediolateral oblique (MLO) views are included
for most of the exams. We treated each view as a separate image in
our experiments.  

Although the dataset has an official train-test split, the body of literature does not follow the suggested split. For a fair comparison, we  randomly split the training data
85:15 at the \emph{patient level} to create independent training and validation
datasets. Overall, there were 2097, 361, 645 mammograms of 1061,
187, 349 women in the training, validation, test data respectively.

To show generalisation capabilities of the transfer learning 
of the transfer capabilities of the proposed approach, we use the 
ChestX-ray14 \cite{WangPengEtAl2017} dataset, which is a hospital-scale database consisting of $112,120$ chest x-ray images with 14 abnormalities. We used the official dataset split.


\subsection{Parameter Selection}
The mammograms in the CBIS-DDSM data are with a mean of $3138\times5220$
pixels. In the proposed experiments, we use random crop and random
horizontal flip to augment data. The random crop ratio is set to be
0.875. For example, an input size of $1792\times2304$ means the mammogram
is resized to $2048\times2633$ with an input image of $1792\times2304$ randomly cropped
to feed into the neural network. The weights of the last fully-connected layer is initialised with the Xavier scheme. For the network training, we follow
the hyper-parameters setup in \cite{ShenMargoliesEtAl2019}. The batch
size was set to be 32, and Adam \cite{KingmaBa2014} was used as the
optimizer. 

Moreover, we follow the 2-stage training strategy in \cite{ShenMargoliesEtAl2019} with the following modifications: i) we set weight decay to its default value $10^{-4}$ because there wasn't a noticeable benefit for changing its value; ii) we did not freeze certain layers in stage-1 because it could slightly
hurt the performance. The following strategy was adopted in the proposed
experiments: Stage-1) Set learning rate to $10^{-4}$ and train all the layers for 30 epochs, and   Stage-2) Set learning rate to $10^{-5}$ and train all the layers for 20 epochs for a total of 50 epochs.

\subsection{Results  and Discussion}
In this section, we detail our findings and give deeper insights into the good performance of the proposed MorphHR approach. 

\begin{table*}
	\begin{centering}
		\begin{tabular}{ccccc}
			\toprule 
			& \cellcolor[HTML]{EFEFEF} Sensitivity  & \cellcolor[HTML]{EFEFEF} Specificity  & \cellcolor[HTML]{EFEFEF} AUC (val)  & \cellcolor[HTML]{EFEFEF} AUC (test)\tabularnewline
			\midrule
			\midrule 
			Single Model Single Crop~\cite{ShenMargoliesEtAl2019}  & -  & -  & 85\%  & $\sim$75\% \tabularnewline
			\midrule 
			Single Model~\cite{ShenMargoliesEtAl2019}   & -  & -  & 88\%  & -\tabularnewline
			\midrule 
			Four Models Ensemble~\cite{ShenMargoliesEtAl2019}  & 86.10\%  & 80.10\%  & 91\%  & -\tabularnewline
			\midrule 
			Single Model Single Crop (MorphHR)  & -  & -  & \cellcolor[HTML]{DCEDC8} 91.82\%  & \cellcolor[HTML]{DCEDC8} 79.64\%\tabularnewline
			\midrule 
			Single Model (MorphHR)  & \cellcolor[HTML]{F1F8E9} 86.98\%  & \cellcolor[HTML]{F1F8E9} 82.57\%  & \cellcolor[HTML]{DCEDC8} 93.48\%  & \cellcolor[HTML]{F1F8E9} 81.87\%\tabularnewline
			\midrule 
			Four Models Ensemble (MorphHR)  & \cellcolor[HTML]{DCEDC8} \textbf{90.0\%}  & \cellcolor[HTML]{DCEDC8} 86.24\%  & \cellcolor[HTML]{DCEDC8} 94.27\%  & \cellcolor[HTML]{F1F8E9} 83.13\%\tabularnewline
			\midrule \midrule 
			Human (Estimation) \cite{LehmanAraoEtAl2016}  & 86.9\%  & \cellcolor[HTML]{F0F4C3} \textbf{88.9\%}  & -  & -\tabularnewline
			\bottomrule
		\end{tabular}
		\par\end{centering}
	\caption{Numerical comparison of the proposed MorpHR scheme, the SOTA-model of that~\cite{ShenMargoliesEtAl2019} and human estimation \cite{LehmanAraoEtAl2016} on the
		CBIS-DDSM dataset. \label{tab:ddsm_compare}}
\end{table*}

\begin{figure}[t!]
	\begin{centering}
		\includegraphics[width=0.90\linewidth]{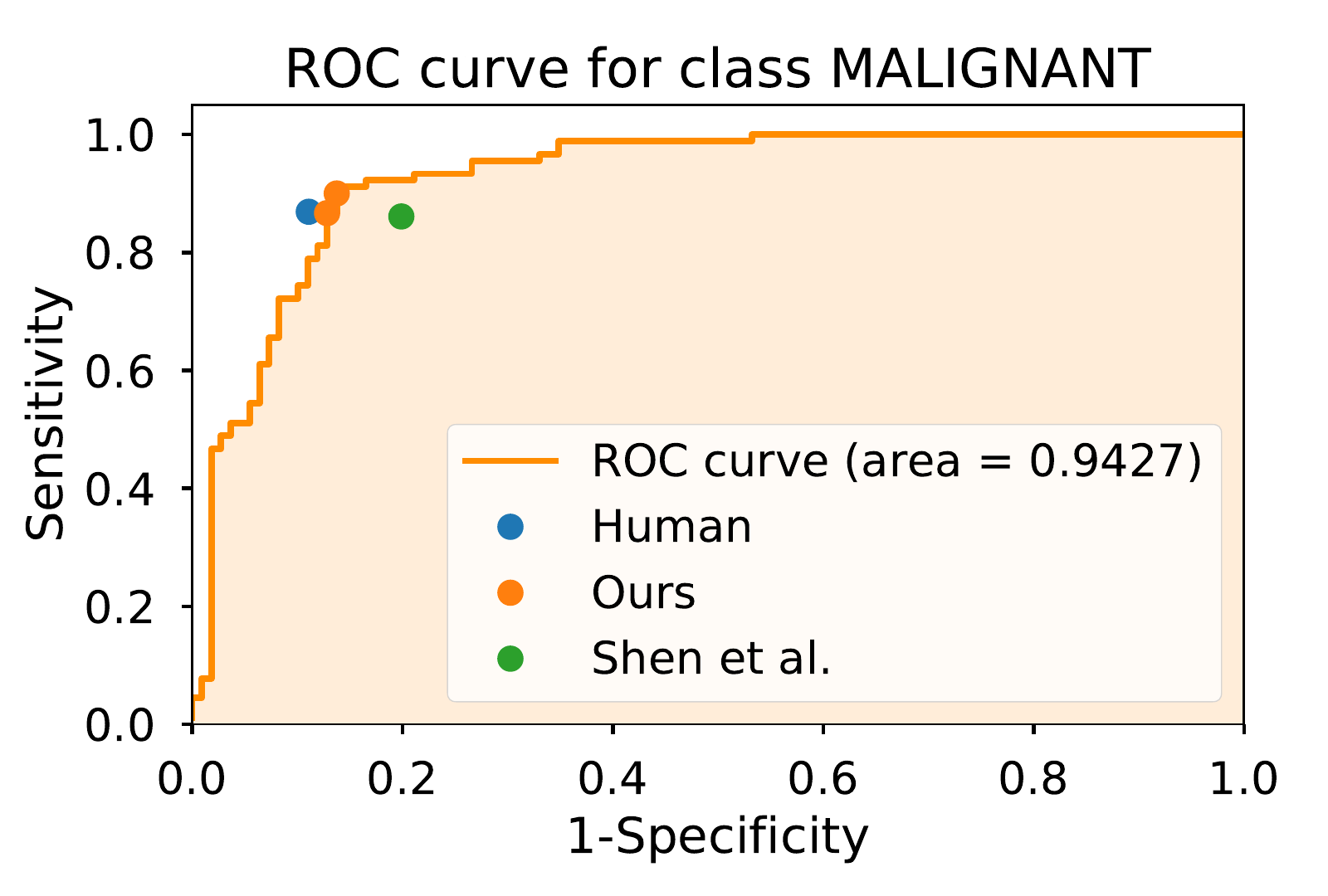} 
		\par\end{centering}
	\caption{Sensitivity and Specificity on ROC curve. The proposed system achieved performance on a par with mammography experts. \label{fig:ddsm_human}}
\end{figure}

\begin{table*}
	\begin{centering}
		\begin{tabular}{ccccc}
			\toprule 
			& \cellcolor[HTML]{EFEFEF} Layer4 computed on & \cellcolor[HTML]{EFEFEF} Input Size & \cellcolor[HTML]{EFEFEF} Test & \cellcolor[HTML]{EFEFEF} Test (10-crops)\tabularnewline
			\midrule
			\midrule 
			ResNet18\_S224 & 7x7 & 224x224 & 79.74 & 81.55\tabularnewline
			\midrule  
			MorphHR-ResNet18\_S224 & 7x7 & 448x448 & \cellcolor[HTML]{DCEDC8} 80.98 & \cellcolor[HTML]{DCEDC8} 82\tabularnewline
			\midrule 
			ResNet18\_S448 & 14x14 & 448x448 & 80.67 & 82.44\tabularnewline
			\midrule 
			MorphHR-ResNet18\_S448 & 14x14 & 896x896 & \cellcolor[HTML]{DCEDC8} 82.53 & \cellcolor[HTML]{DCEDC8} \textbf{83.77}\tabularnewline
			\bottomrule
		\end{tabular}
		\par\end{centering}
	\caption{Experimental results of the proposed MorphHR scheme on the
		ChestX-ray14 (Mass) dataset. \label{tab:chest_results_mass}}
\end{table*}

\begin{table*}
	\begin{centering}
		\resizebox{\textwidth}{!}{ 
			\begin{tabular}{ccccccccc}
				\toprule 
				& \cellcolor[HTML]{EFEFEF} In 8 & \cellcolor[HTML]{EFEFEF} Support & \cellcolor[HTML]{EFEFEF} Wang \cite{WangPengEtAl2017} & \cellcolor[HTML]{EFEFEF} Wang \cite{YaoProskyEtAl2018} & \cellcolor[HTML]{EFEFEF} Yao \cite{YaoProskyEtAl2018} & \cellcolor[HTML]{EFEFEF} GraphXNet \cite{Aviles-RiveroPapadakisEtAl2019} & \cellcolor[HTML]{EFEFEF} MorphHR & \cellcolor[HTML]{EFEFEF} MorphHR (10-crops)\tabularnewline
				\midrule
				\midrule 
				Atelectasis & T & 11559 & 70.69 & 70.03 & 73.3 & 71.89 & \cellcolor[HTML]{DCEDC8} 76.12 & \cellcolor[HTML]{DCEDC8} 77.62\tabularnewline
				\midrule 
				Cardiomegaly & T & 2776 & 81.41 & 81 & 85.6 & \cellcolor[HTML]{DCEDC8} 87.99 & 85.81 & 86.53\tabularnewline
				\midrule 
				Consolidation &  & 4667 & - & 70.32 & 71.1 & 73.36 & \cellcolor[HTML]{DCEDC8} 77.38 & \cellcolor[HTML]{DCEDC8} 77.60\tabularnewline
				\midrule 
				Edema &  & 2303 & - & 80.52 & 80.6 & 80.20 & \cellcolor[HTML]{DCEDC8} 84.77 & \cellcolor[HTML]{DCEDC8} 85.44\tabularnewline
				\midrule 
				Effusion & T & 13317 & 73.62 & 75.85 & 80.6 & 79.20 & \cellcolor[HTML]{DCEDC8} 81.50 & \cellcolor[HTML]{DCEDC8} 82.97\tabularnewline
				\midrule 
				Emphysema &  & 2516 & - & 83.3 & 84.2 & 84.07 & \cellcolor[HTML]{DCEDC8} 88.98 & \cellcolor[HTML]{DCEDC8} 89.78\tabularnewline
				\midrule 
				Fibrosis &  & 1686 & - & 78.59 & 74.3 & 80.34 & \cellcolor[HTML]{DCEDC8} 80.59 &  \cellcolor[HTML]{DCEDC8} 81.36\tabularnewline
				\midrule 
				Hernia &  & 227 & - & \cellcolor[HTML]{F1F8E9} 87.17 & 77.5 & \cellcolor[HTML]{F1F8E9} 87.22 & 85.42 & \cellcolor[HTML]{DCEDC8} 88.22\tabularnewline
				\midrule 
				Infiltration & T & 19894 & 61.28 & 66.14 & 67.3 & \cellcolor[HTML]{F1F8E9} 72.05 & 71.13 & \cellcolor[HTML]{DCEDC8} 72.42\tabularnewline
				\midrule 
				Mass & T & 5782 & 56.09 & 69.33 & 77.7 & 80.90 & \cellcolor[HTML]{DCEDC8} 82.53 & \cellcolor[HTML]{DCEDC8} 83.77\tabularnewline
				\midrule 
				Nodule & T & 6331 & 71.64 & 66.87 & 71.8 & 71.13 & \cellcolor[HTML]{DCEDC8} 75.15 & \cellcolor[HTML]{DCEDC8} 77.07\tabularnewline
				\midrule 
				Pleural\_Thickening &  & 3385 & - & 68.35 & 72.4 & 75.70 & \cellcolor[HTML]{DCEDC8} 78.44 & \cellcolor[HTML]{DCEDC8} 80.00\tabularnewline
				\midrule 
				Pneumonia & T & 1431 & 63.33 & 65.8 & 68.4 & \cellcolor[HTML]{DCEDC8} 76.64 & 69.70 & 72.11\tabularnewline
				\midrule 
				Pneumothorax & T & 5302 & 78.91 & 79.93 & 80.5 & 83.70 & \cellcolor[HTML]{DCEDC8} 86.52 & \cellcolor[HTML]{DCEDC8} 87.01\tabularnewline
				\midrule
				\midrule 
				Average AUC &  & - & - & 73.8 & 76.1 & 78.88 & \cellcolor[HTML]{DCEDC8} 80.29 & \cellcolor[HTML]{DCEDC8} \textbf{81.56}\tabularnewline
				\bottomrule
			\end{tabular}
		}
		\par\end{centering}
	\caption{Numerical comparison of the proposed MorpHR scheme and the SOTA-model on the
		ChestX-ray14 dataset. \label{tab:chest_results}}
\end{table*}
\begin{figure*}[h!]
	\begin{centering}
		\hspace*{-0.25in}\subfloat[\label{fig:chest_radar}]{\begin{centering}
				\includegraphics[width=0.40\linewidth]{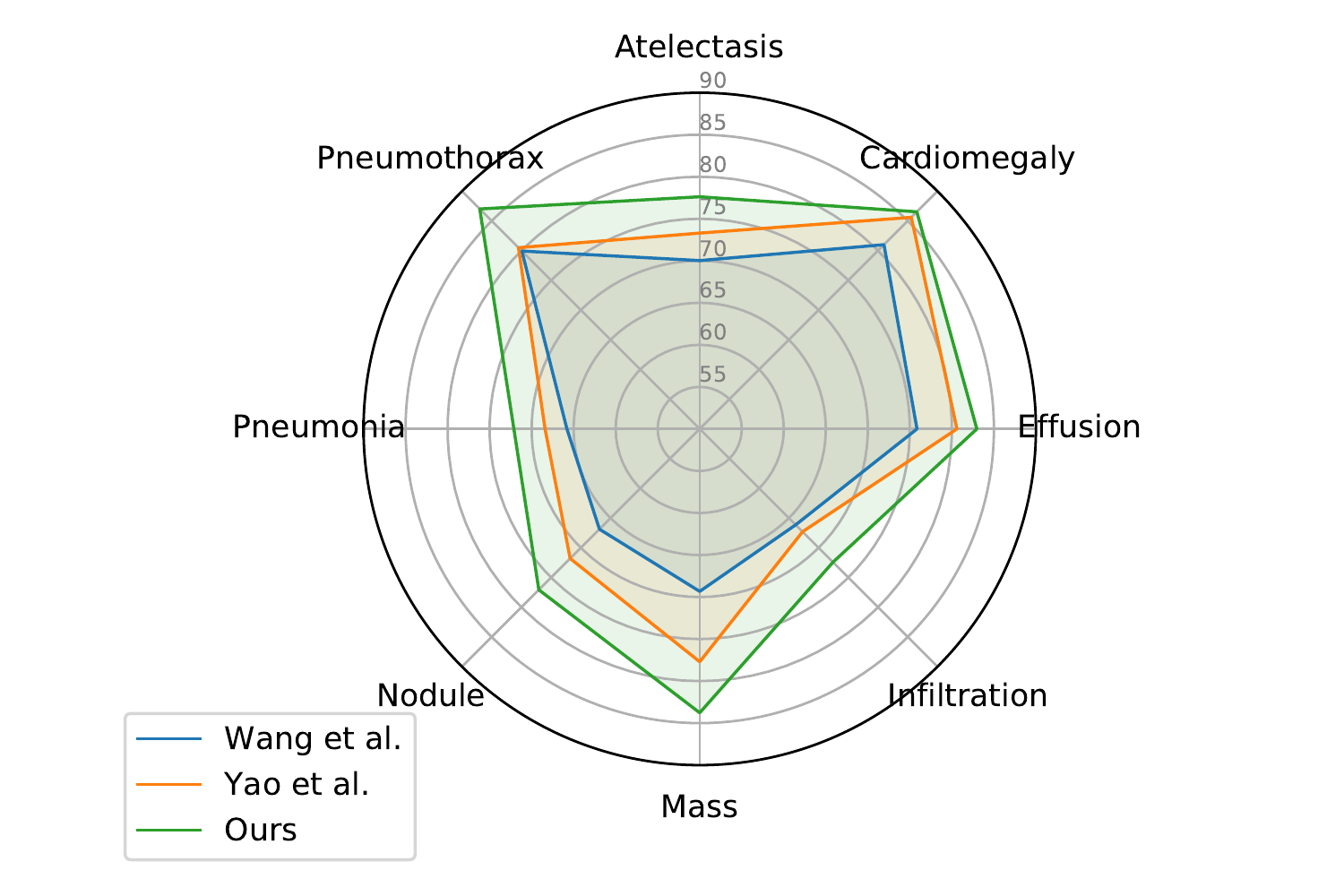} 
				\par\end{centering}
		}\hspace*{-0.35in}\subfloat[\label{fig:chest_roc_curves}]{\begin{centering}
				\includegraphics[width=0.29\linewidth]{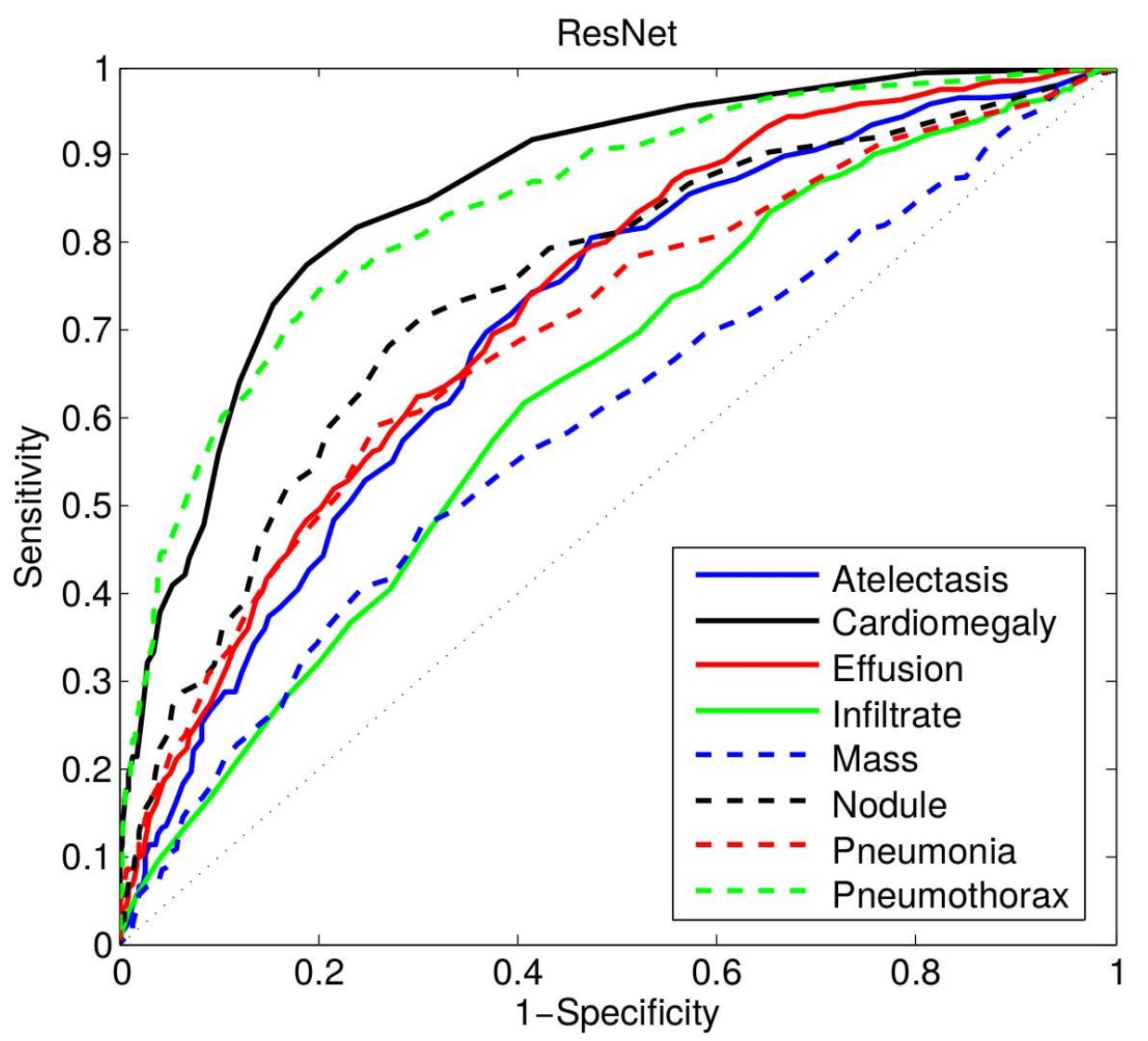}
				\includegraphics[width=0.33\linewidth]{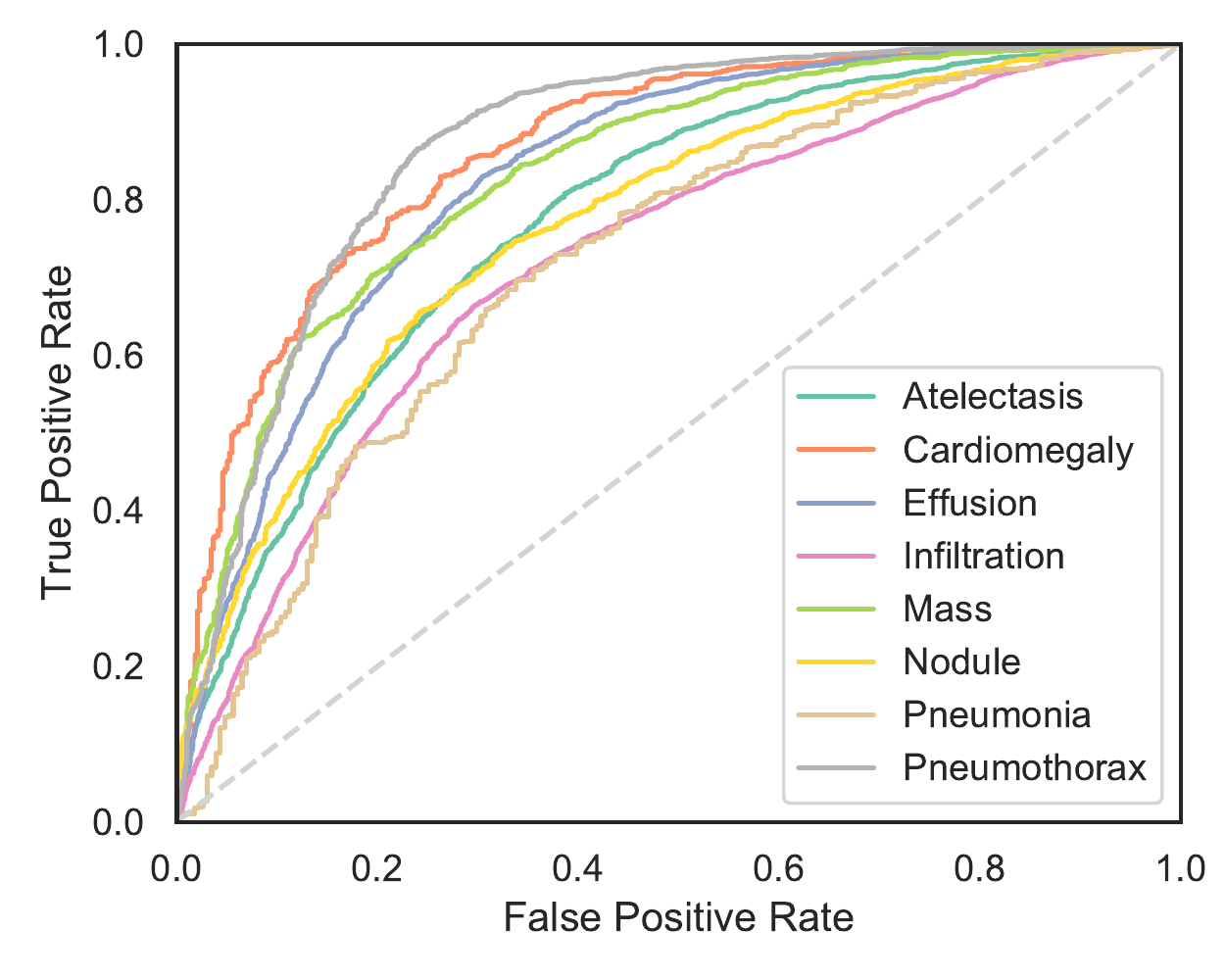}
				\par\end{centering}
		}
		\par\end{centering}
	\caption{Comparison of the proposed MorpHR scheme against existing approaches in (a) Radar plot (\cite{WangPengEtAl2017} and \cite{YaoProskyEtAl2018}).
		(b) Left: ROC curve from \cite{WangPengEtAl2017}; right: ROC curve of the proposed approach. }
\end{figure*}
\smallskip

\textbf{Fine-Tuning vs MorphHR.} We begin our evaluation by comparing the performance of the proposed approach against fine-tuning using a ROC-AUC analysis. We first compare the performances of ResNet and MorphHR-ResNet, on the  CBIS-DDSM dataset, which results are displayed in Table~\ref{tab:ddsm_results}.
In this table, we show the AUC scores on the validation and test data, with both single-crop and 10-crops results.  In a closer look at the results, one can see that the proposed MorphHR scheme has a consistent improvement over the baseline counterpart. 
The performance improvement on the validation and testing data can be up to 4.35\% (82.5\% vs 78.15\%) and 2.66\% (75.23\% vs 72.57\%) respectively.
We remark that ``Ensemble'' denotes an average of the output scores run in four rounds, we include it in order to compare against the ensemble of fours models of that~\cite{ShenMargoliesEtAl2019}.

\textcolor{black}{
\textbf{Resolution is Critical for High Resolved Screening Mammogram Classification.} We reveal that resolution is critical for mammogram classification. It is intuitive that mammograms are high resolved ($\sim$4K) in nature. However, due to the limit of GPUs memory, current mainstream approaches either decrease the resolution or to extract patches. We argue that the importance of resolution for medical image classification should be emphasised. In particular, this research is proposed to resolve this resolution issue. }

To further support the results, we display the ROC curves of fine-tuning vs MorphHR using different input sizes. The results are reported in Fig.~\ref{fig:ddsm_roc_curve}, they summarise the performances. From a comparison at those plots, one can observe that the proposed approach performs better than fine-tuning at all thresholds. Overall, one can observe that
\textit{the image resolution plays a very important role in the task of screening mammography classification.} This is meaningful because a higher resolution image gives more fine details, especially when considering the fact that the lesion regions only accounts for a small part of the whole image.

\medskip
\textbf{MorphHR vs SOTA-Model.} We now compare the proposed approach against the SOTA method of that~\cite{ShenMargoliesEtAl2019}, which to the best of our knowledge holds the state-of-the-art results in the   CBIS-DDSM dataset. The results are displayed in  Table \ref{tab:ddsm_compare}.

On the validation dataset, using four models ensemble, we improve current
best results of 91\% in \cite{ShenMargoliesEtAl2019} to 94.27\%.
In summary, there is a 7\% improvement for single model single crop, a 5.5\% improvement
for single model, and a 3\% improvement for four models ensemble. The sensitivity
and specificity are also improved from 86.10\% and 80.10\% to 90.00\% and 
86.24\% respectively. 
The sensitivity and specificity of screening
mammography of human radiologists are reported to be an average of 86.9\% and 88.9\% respectively \cite{LehmanAraoEtAl2016}.
It can be seen that the proposed CAD system is on par with mammography experts. This is further illustrated in Fig. \ref{fig:ddsm_human}, where the sensitivity and specificity pairs on the ROC curves are drawn.  

In Table \ref{tab:ddsm_compare},
we also include the experimental results on the testing dataset. In
\cite{ShenMargoliesEtAl2019}, the authors used a custom split and
did not report results on the official testing dataset. The 75\% AUC
score is adopted from \cite{Shen2019GithubIssue5}, 
which is reported on the official testing data using the original
authors' deployed models. It can be seen on the testing dataset, we
are able to achieve a 4.64\% AUC improvement (75\% vs 79.64\%).
One may notice that in Table \ref{tab:ddsm_compare}, for both \cite{ShenMargoliesEtAl2019} and the proposed MorphHR scheme, the performances on the testing data is significantly lower than those on the validation data. 
This is because the testing data is another holdout set acquired in a different time. It contains cases which are intrinsically more difficult and bear different distributions \cite{Shen2019GithubIssue5}.

\textbf{Generalisation Capabilities: The X-ray Case.} To demonstrate the generalisation capability of the proposed approach for transferring knowledge, we further carry out experiments on the ChestX-ray14 \cite{WangPengEtAl2017} dataset. We follow the same experimental protocol as the previous section. 

\emph{Fine-Tuning vs MorphHR}. We start by first supporting our claim regarding the limitation of fine-tuning for medical data. 
In Table~\ref{tab:chest_results_mass}, we report a performance comparison in terms of AUC,  the results reflect the outputs on the ``Mass" class, which is close related to the mammography case. 
From these results, one can observe that the proposed approach report the best performance - that is, the results are consistent with those findings on the CBIS-DDSM dataset. By using the proposed MorphHR scheme, we can achieve up to 2\% (82.53\% vs 80.67\%) AUC performance improvement than using only fine-tuning. Hence, the effectiveness of the proposed MorphHR scheme is demonstrated.

%


\emph{MorphHR vs SOTA-Model.} To further support of our results, we also reported the average AUC scores on all pathologies, the results are reported in Table \ref{tab:chest_results}. In a detail inspection, one can observe that overall the proposed approach reports SOTA results for this dataset. We can also observe that there are only few pathologies with clear variability. It is because of the limitation of the dataset, for example  ``Hernia" is the one with the fewest samples in the dataset. The dataset, therefore, is not always representative for each class (despite the data augmentation).

We remark that other results reported on this dataset are, either not
on the official test data \cite{GuendelGrbicEtAl2018,BaltruschatNickischEtAl2019},
or using per-image split \cite{YaoPoblenzEtAl2017,RajpurkarIrvinEtAl2017},
whose training and testing data can duplicate because each patient
has two images of each breast. Fig. \ref{fig:chest_radar} compares
against \cite{WangPengEtAl2017,YaoProskyEtAl2018} in radar plot and
Fig. \ref{fig:chest_roc_curves} compares against \cite{WangPengEtAl2017} in ROC curves. The advantage of the proposed MorphHR scheme is obvious. 


Overall, to the best of our knowledge, we are reporting SOTA performances for both medical dataset cases -  that is, the mammography and X-ray data classification.


\section{Conclusions and Future Work}
In this work, we proposed a new transfer learning framework for improving mammography data classification beyond fine-tuning, \textcolor{black}{ which learns domain specific features, accepts high resolution inputs and is scalable to hardware. }
Fine-tuning is the de-facto method for deep learning application to
medical imaging. However, it only allows the modification of the last
few layers of a neural network to adapt the high-level concept class
information into the new dataset. In this research, we proposed to use
network morphism to alter the front of a neural network to learn the
considerable differences between natural images and medical images.
In particular, we also proposed a concrete learning scheme to deal with
the high resolution nature of mammographic images. Extensive experiments
were carried out on the benchmark datasets CBIS-DDSM and ChestX-ray14
to achieve state-of-the-art results.

In this research, the proposed modification of the standard ImageNet models for high resolution mammogram classification is simple and effective. It is meaningful we can design more elegant patterns for block $A'$ in Fig. \ref{fig:illustration} to further improve the performance, which will be the focus for future work. 

\section*{Acknowledgment}
AIAR gratefully acknowledges the financial support of the CMIH and CCIMI University of Cambridge.
CBS acknowledges support from the Leverhulme Trust project on Breaking the non-convexity barrier, the Philip Leverhulme Prize 2018, EPSRC grant Nr. EP/M00483X/1, the EPSRC Centre Nr. EP/N014588/1, the RISE projects CHiPS and NoMADS, the Cantab Capital Institute for the Mathematics of Information and the Alan Turing Institute. FJG is an NIHR Senior Investigator.

{\bibliographystyle{ieeetr}
	

}


%

\end{document}